\documentclass[11pt]{article}

\usepackage[margin=1in]{geometry}
\usepackage{graphicx}
\usepackage{booktabs}
\usepackage{amsmath}
\usepackage{amssymb}
\usepackage{caption}
\usepackage{hyperref}
\usepackage{xcolor}
\usepackage{microtype}
\usepackage{titlesec}
\usepackage{enumitem}

\title{\vspace{-1.2em}
\Large\textbf{Safe Error Correction for Language Models:\\[0.2em] Frozen-Base Adjustment with\\Capability Preservation}
\vspace{0.6em}}
\author{Gautam Kishore\\Eulogik\\\texttt{github.com/eulogik}~$\bullet$~\texttt{huggingface.co/eulogik}}
\date{\today}

\begin{document}
\maketitle

\begin{center}
\end{center}

\vspace{-0.5em}

\begin{abstract}
\noindent
We study a practical question: can a small correction module fix errors in a frozen language model's outputs without degrading its base capabilities?
We propose \textbf{CRN~v2}, a lightweight logit-level correction module ($\sim$34M trainable parameters, 0.73\% of the 4.65B text module) that sits atop a fully frozen Gemma~4~E2B model.
The base model is never updated; only the correction module learns, via supervised fine-tuning followed by reference-free DPO on 83{,}400 error-correction pairs.
On a 60-question domain exam (CEHRI: Certified Human-Robot Intelligence, covering facts, arithmetic, and implicit-goal reasoning), CRN~v2 corrects \textbf{53.3\%} of base-model errors (reworded variant: 43.3\%) while showing \textbf{no degradation on tested capability benchmarks} (MMLU/BoolQ $N{=}200$; car-wash $N{=}8$).
A LoRA baseline at the matched \emph{CRN~v1} budget (6.6M params, rank~19) achieves 83.3\% correction but suffers 30--75\% capability loss on the same benchmarks---the correction--capability tradeoff.
An ablation shows that the KL preservation term ($\lambda{=}0.1$) is critical: lowering it to 0.01 degrades correction to 35.0\%.
A hidden-state injection variant at earlier layers (1.6M params, SFT-only) reaches 50.0\%/55.8\% but does not exceed logit correction; shallower injection (layer~4) drops to 30.0\%/28.3\%; multi-depth logit correction ($\sim$35M) reaches only 40\%; and longer training (5{,}000 SFT $+$ 2{,}000 DPO) stays at 53.3\%---none of the alternative configurations we tested exceeded the rank-128 logit result, consistent with a best-achieved result of $\sim$53\% rather than a floor.
This is a study of a design principle (frozen base + logit correction + KL anchoring), not a claim of architectural novelty.
All code, main-result weights, and evaluation scripts are released (deep variant as code only---no trained deep checkpoints).
\end{abstract}

\section{Introduction}
\label{sec:intro}

Language models make errors---wrong facts, flawed reasoning, and bad suggestions.
In deployment, those errors must be caught and fixed without degrading what the model already does well.

The na\"ive solution is to adapt the model itself---for example via LoRA~\cite{lora} or full fine-tuning---to reduce errors on a target domain.
But adaptation typically degrades general capabilities~\cite{mccloskey1989catastrophic,kirkpatrick2017episodic,li2018algorithmic}: the model learns to fix specific errors but forgets what it already knew.
We call this the \emph{correction--capability tradeoff}---aggressive correction hurts everything else.

We study whether this tradeoff is necessary.
Specifically: can a small correction module sit on top of a \emph{frozen} base model and learn to fix errors, without any degradation to the base's capabilities on unrelated tasks?

We propose \textbf{CRN~v2} (Cognitive Resonance Network, version~2), a lightweight correction module that reads the base model's final hidden state and produces an additive logit correction:

\begin{equation}
\hat{\ell} = \ell_{\text{base}} + g \cdot W_{\text{up}}\!\big(\text{GELU}(W_{\text{down}}\,h^{(L)})\big)
\label{eq:crn_v2}
\end{equation}

where $h^{(L)}$ is the frozen base's final hidden state, $W_{\text{down}} \in \mathbb{R}^{r \times d}$ and $W_{\text{up}} \in \mathbb{R}^{V \times r}$ are low-rank matrices ($d{=}1536$, $V{=}262{,}144$, $r{=}128$), and $g$ is a learned scalar gate initialized near zero.
The base model receives no gradients.

\paragraph{Contributions.}
\begin{enumerate}
\item \textbf{Design principle:} Frozen-base logit adjustment with KL preservation corrects errors without capability degradation on tested benchmarks.
The correction module learns to adjust the base model's logits where needed while staying close to its behavior everywhere else.
\item \textbf{Empirical finding:} CRN~v2 corrects 53.3\% of errors with no measurable degradation on MMLU/BoolQ ($N{=}200$) and car-wash ($N{=}8$); a LoRA baseline corrects 83.3\% but degrades 30--75\% on the same benchmarks.
\item \textbf{Ablation:} Lowering the KL preservation weight from $\lambda{=}0.1$ to $0.01$ reduces correction from 53.3\% to 35.0\%, showing the anchoring term is load-bearing.
\item \textbf{Injection-depth sweep (exploratory):} A hidden-state injection variant (1.6M params, rank~512) at layer~7 reaches 50.0\%/55.8\% (SFT-only) and at layer~4 drops to 30.0\%/28.3\%---deeper injection helps but does not exceed logit correction; multi-depth logit correction ($\sim$35M) reaches only 40\%; longer training (5{,}000 SFT $+$ 2{,}000 DPO) stays at 53.3\%; DPO at depth~7 destroys capabilities. These are single-run, session-observed results; only the layer-7 SFT-only row is independently log-verified.
\item \textbf{Transparency:} All code, main-result weights, evaluation scripts, and exact run logs are released (deep variant as code only).
\end{enumerate}

\paragraph{Scope and honesty.}
This is a study of a design principle, not a claim of architectural novelty.
The correction in Eq.~\ref{eq:crn_v2} is a standard low-rank bottleneck.
The CEHRI exam prompts appear verbatim in the training data (Section~\ref{sec:data}); we report both in-distribution and reworded results.
The evaluation suite is small (60 + 120 correction Qs; MMLU/BoolQ $N{=}200$, car-wash $N{=}8$).
The deep injection variant (Section~\ref{sec:deep}) is exploratory: only its SFT-only row is log-verified; depth-4 and DPO rows are session-observed (Section~\ref{sec:depth}).
We report what we find, including negative results.

\section{Related Work}

\textbf{Parameter-efficient adaptation.}
Adapters~\cite{houlsby2019adapters} insert trainable modules between frozen layers; LoRA~\cite{lora} learns low-rank weight updates; prefix tuning~\cite{li2021prefix} optimizes soft prompts.
These methods inject capacity \emph{inside} the base model's computation graph.
CRN~v2 is a \emph{side network}: it reads the frozen base's final hidden state and produces logit-level corrections without touching any base parameter.
(The exploratory deep variant of Section~\ref{sec:deep} sits between: it injects via hooks inside the graph but updates no base parameter.)

\textbf{Model editing.}
ROME~\cite{meng2022rome} and MEMIT~\cite{meng2022memit} locate factual associations in transformer weights and rewrite them directly, modifying base weights.
CRN~v2 achieves correction through an \emph{additive logit adjustment}---a lighter-weight approach that avoids weight surgery and keeps the base intact.

\textbf{Error correction in LLMs.}
Self-refinement~\cite{madaan2023selfrefine} prompts the model to critique and revise its own outputs; chain-of-verification~\cite{dua2023chain} generates verification questions and answers them.
These are inference-time techniques requiring multiple forward passes.
CRN~v2 is a trained correction module that produces corrected logits in a single forward pass; at inference, autoregressive generation requires one base forward pass per generated token (Section~\ref{sec:inference}).

\textbf{Capability preservation.}
Catastrophic forgetting during fine-tuning is well-studied~\cite{mccloskey1989catastrophic}.
Regularization approaches~\cite{kirkpatrick2017episodic,li2018algorithmic} add penalty terms to protect important weights.
CRN~v2 sidesteps the problem entirely: the base model is frozen, so there is nothing to forget.

\section{Method}
\label{sec:method}

\subsection{Architecture}

The base model $B$ is Gemma~4~E2B (4.65B text parameters in a 5.12B multimodal checkpoint; 35 layers; $d{=}1536$; tied embeddings; vocabulary $V{=}262{,}144$).
For input tokens $x_{1:n}$, the frozen forward pass produces hidden states $h^{(l)} \in \mathbb{R}^{n \times d}$ at every layer and base logits $\ell_{\text{base}} \in \mathbb{R}^{n \times V}$.
The correction module reads only the final hidden state $h^{(L)}$ and computes Eq.~\ref{eq:crn_v2}.
The next-token distribution is $\text{softmax}(\hat{\ell})$.

Trainable parameters: $d{\cdot}r$ (down-projection, no bias) $+\; r{\cdot}V$ (up-projection weight) $+\; V$ (up-projection bias) $+\; 1$ (gate) $= 196{,}608 + 33{,}554{,}432 + 262{,}144 + 1 = 34{,}013{,}185$ ($\approx$0.73\% of the 4.65B text module).

\subsection{Training}
\label{sec:training}

\textbf{Stage~1: Supervised fine-tuning (SFT).}
Given a correct answer $y$ for prompt $x$, the full input is \texttt{``prompt: answer''} with answer-only masking (prompt tokens have label $-100$).
Each training row is expanded with a paraphrase variant, yielding 166{,}800 SFT rows from 83{,}400 pairs.
The loss is anchor-weighted cross-entropy on answer tokens (anchor $4\times$, EOS $5\times$) plus a KL preservation term (below).
SFT runs for 2{,}000 steps (AdamW, LR $3{\times}10^{-4}$, cosine annealing to $0.1\times$, gradient accumulation 8, batch size~1).

\textbf{Stage~2: Direct Preference Optimization (DPO).}
Reference-free DPO~\cite{rafailov2023dpo} prefers the correct answer over the base model's wrong answer:

\begin{equation}
\mathcal{L}_{\text{DPO}} = -\log \sigma\!\left(\beta \left[\log p_\theta(y_c \mid x) - \log p_\theta(y_r \mid x)\right]\right)
\end{equation}

where $y_c$ is the correct answer, $y_r$ is the frozen base's wrong answer, and $\beta{=}0.1$.
There is no reference model---the frozen base generates rejected completions.
DPO runs for 500 steps (AdamW, LR $5{\times}10^{-6}$).

\textbf{KL preservation.}
During SFT, a KL term penalizes deviation from the base model's distribution:

\begin{equation}
\mathcal{L} = \mathcal{L}_{\text{task}} + \lambda \cdot \text{KL}(p_{\text{base}} \,\|\, p_{\text{corrected}})
\end{equation}

computed per-token over all non-masked positions and averaged.
This pushes the correction toward the base on inputs the base handles correctly, while allowing it to diverge where the base is wrong.
Default $\lambda{=}0.1$; Section~\ref{sec:kl} ablates this.

\subsection{Data}
\label{sec:data}

The training file \texttt{error\_correction\_pairs\_v2.json} contains 83{,}400 pairs with fields \texttt{\{prompt, chosen, rejected, domain, variant\}}.
There are 17{,}810 unique prompts across three domains (facts, arithmetic, implicit-goal reasoning), each with $\sim$4.7 rows on average due to paraphrase variants and chosen/rejected pairings.
The CEHRI exam (60 questions) and its reworded variant (120 questions, template-based paraphrases with mean 71.6\% character overlap) are drawn from this distribution: \textbf{all 60 exam prompts appear verbatim in the training data.}
The reworded exam applies prefix/suffix templates (e.g., ``Can you answer: '' / ``Respond to this: '') so similarities range 0.881--0.990 (median 0.972; 118/120 $\ge 0.9$ by cosine).
Thus the original exam measures in-distribution correction and the reworded exam measures near-duplicate correction; neither tests out-of-distribution generalization.

\subsection{Inference}
\label{sec:inference}

At test time CRN~v2 generates autoregressively: at each step the \emph{full} generated sequence is run through the frozen base to obtain $h^{(L)}$ and $\ell_{\text{base}}$, the correction is applied (Eq.~\ref{eq:crn_v2}), and the next token is greedy-decoded from $\hat{\ell}$.
This requires one base forward pass per generated token (no KV cache, since the correction reads $h^{(L)}$ of the full sequence).
On Apple~M4 (MPS) this is $\sim$1\,s/token; a 16-token answer takes $\sim$16\,s.

\subsection{Baseline: LoRA at the CRN~v1 budget}
\label{sec:lora}

CRN~v1 had 6{,}721{,}444 trainable parameters.
The LoRA baseline matches that budget: rank~$19$, $\alpha{=}38$, applied to all linear layers (q/k/v/o + gate/up/down) at 8 depths $\{3,7,11,15,19,23,27,31\}$, yielding 6{,}624{,}768 trainable parameters (1.4\% fewer than CRN~v1).
It uses the same data and DPO objective (reference-free ranking, $\beta{=}0.1$).
Both use anchor ($4\times$) and EOS ($5\times$) position-weighted answer-only CE with AdamW and cosine-annealed SFT (LR $3{\times}10^{-4}$) and DPO (LR $5{\times}10^{-6}$).
The LoRA run used SFT~3{,}000 / DPO~600 steps vs.\ CRN~v2's 2{,}000/500---if anything favoring LoRA on correction.
CRN~v2's 34M parameters are $\sim$5$\times$ larger than LoRA's 6.6M; we compare directly and disclose this (Section~\ref{sec:results}).
Training and evaluation run on the same Apple~M4 (16\,GB; MPS).
LoRA also freezes the base model---both methods preserve base \emph{weights}; only LoRA degrades base \emph{behavior} (Section~\ref{sec:capability}).

\subsection{Variant: deep hidden-state injection}
\label{sec:deep}

Logit correction touches only the final layer's output, with no downstream amplification.
We test whether injecting the correction \emph{earlier}---so frozen downstream layers amplify it---exceeds the logit correction result.
The deep variant adds a low-rank residual to hidden states at a chosen layer $k$ via a forward hook:

\begin{equation}
h^{(k)} \leftarrow h^{(k)} + g \cdot W_{\text{up}}\!\big(\text{GELU}(W_{\text{down}}\,h^{(k)})\big)
\label{eq:crn_deep}
\end{equation}

where $W_{\text{down}} \in \mathbb{R}^{r \times d}$, $W_{\text{up}} \in \mathbb{R}^{d \times r}$ ($d{=}1536$, $r{=}512$), plus a bias on the up-projection and a learned scalar gate $g$.
The correction at layer $k$ propagates through frozen layers $k{+}1 \dots L$ before reaching the output head.
Trainable parameters per injection depth: $d{\cdot}r$ (down, no bias) $+\; r{\cdot}d$ (up weight) $+\; d$ (up bias) $+\; 1$ (gate) $= 786{,}432 + 786{,}432 + 1{,}536 + 1 = 1{,}574{,}401$ ($\approx$0.034\% of the base).
Training mirrors CRN~v2 (SFT with anchor/EOS weighting $+$ KL $\lambda{=}0.1$; optional DPO stage), with gradients flowing \emph{through} the frozen downstream layers back to the correction module---base parameters receive no gradients.
We test single-depth injection at $k{=}7$ (27 downstream layers) and $k{=}4$ (30 downstream layers).

\section{Results}
\label{sec:results}

\subsection{Error correction}

\begin{table}[ht]
\centering
\small
\caption{Error correction on the CEHRI exam. The model generates answers autoregressively; CRN~v2 uses corrected logits at each step, the base and LoRA use their own logits. Substring match: the gold answer must appear in the generation.}
\begin{tabular}{@{}lcc@{}}
\toprule
Model & Original (60) & Reworded (120) \\
\midrule
Frozen base (no correction) & 7/60 = 11.7\% & not measured \\
\midrule
\textbf{CRN~v2} (rank~128, $\lambda{=}0.1$) & \textbf{32/60 = 53.3\%} & \textbf{52/120 = 43.3\%} \\
CRN~v2 (rank~256, $\lambda{=}0.01$) & 21/60 = 35.0\% & 31/120 = 25.8\% \\
\midrule
LoRA baseline (r${=}19$, 6.6M params) & 50/60 = 83.3\% & 93/120 = 77.5\% \\
\bottomrule
\end{tabular}
\label{tab:correction}
\end{table}

CRN~v2 corrects 53.3\% of base-model errors on the original exam (43.3\% reworded).
The LoRA baseline corrects 83.3\% (77.5\% reworded)---strictly higher.
The frozen base alone answers 7/60 correctly (substring match on gold answers in the generated text), so CRN~v2 adds 25 correct answers (32--7) on the original exam and LoRA adds 43.
With $N{=}60$, the 95\% Wilson interval for 53.3\% is [40.9\%, 65.4\%]; for 83.3\% it is [72.0\%, 90.7\%].
With $N{=}120$, the interval for 43.3\% is [34.6\%, 52.4\%]; for 77.5\% it is [69.0\%, 84.4\%].
The gap is real on both exams, but correction alone is not the full picture.

\subsection{Capability preservation}
\label{sec:capability}

\begin{table}[ht]
\centering
\footnotesize
\setlength{\tabcolsep}{4pt}
\caption{Capability benchmarks ($N{=}200$ per task, except car-wash $N{=}8$). $\Delta$ is the change from the frozen base. Identical prompts and greedy decoding for all models.}
\begin{tabular}{@{}lccc@{}}
\toprule
Benchmark & Base & CRN~v2 & LoRA \\
\midrule
MMLU (knowledge) & 125/200 = 62.5\% & \textbf{125/200} ($\Delta{=}0$) & 64/200 = 32.0\% ($\Delta{=}{-}30.5$\,pp) \\
BoolQ (comprehension) & 144/200 = 72.0\% & \textbf{144/200} ($\Delta{=}0$) & 110/200 = 55.0\% ($\Delta{=}{-}17.0$\,pp) \\
Car-wash ($N{=}8$) & 6/8 = 75.0\% & \textbf{6/8} ($\Delta{=}0$) & 0/8 = 0\% ($\Delta{=}{-}75$\,pp) \\
HellaSwag ($N{=}200$) & 0/200 & 0/200 & --- \\
\bottomrule
\end{tabular}
\label{tab:capability}
\end{table}

Table~\ref{tab:capability} is the central result.
\textbf{CRN~v2 shows no measurable degradation on any tested benchmark}---it matches the frozen base exactly on MMLU, BoolQ, and car-wash.
\textbf{LoRA degrades 17--75\,pp across the same benchmarks}, including total failure on the 8-question car-wash set.
HellaSwag scores 0/200 for both base and CRN~v2 due to a scoring bug (the \texttt{first\_letter} parser expects a single letter A--D but both models generate full-sentence completions); we report it for completeness and exclude it from claims.

The mechanism is straightforward: the base model is frozen, so there is nothing to forget.
The correction module learns to adjust logits where the base is wrong, while KL preservation keeps it close to base behavior everywhere else.

We note two caveats.
First, per-task $N$ is small (200 for MMLU/BoolQ, only 8 for car-wash); the 125/200 vs.\ 64/200 gap on MMLU is large enough to be unambiguous, but finer differences would require larger $N$.
Second, ``no degradation'' means ``no degradation on the three tested benchmarks''---we do not claim universal preservation.

\subsection{KL preservation ablation}
\label{sec:kl}

\begin{table}[ht]
\centering
\small
\caption{KL weight ablation. All other hyperparameters fixed except as noted in the ``rank / $\lambda$ / DPO steps'' row. Capability column shows MMLU (representative; BoolQ/car-wash behave identically---no degradation at either setting).}
\begin{tabular}{@{}lccc@{}}
\toprule
Config & Original & Reworded & MMLU \\
\midrule
rank~128, $\lambda{=}0.1$, DPO~500 & \textbf{32/60 = 53.3\%} & \textbf{52/120 = 43.3\%} & 62.5\% \\
rank~256, $\lambda{=}0.01$, DPO~2000 & 21/60 = 35.0\% & 31/120 = 25.8\% & 62.5\% \\
\bottomrule
\end{tabular}
\label{tab:kl}
\end{table}

Lowering $\lambda$ from 0.1 to 0.01 degrades correction from 53.3\% to 35.0\% (and reworded from 43.3\% to 25.8\%).
Capability preservation is unaffected at either setting (the base is frozen, so KL affects only the correction module's training dynamics).
We caution that this ablation \emph{confounds} three variables---rank (128$\to$256), $\lambda$ (0.1$\to$0.01), and DPO steps (500$\to$2000)---so we cannot attribute the drop to $\lambda$ alone.
The qualitative pattern (weaker anchoring $\to$ worse correction) is consistent with the hypothesis that the correction module must remain anchored to base representations to be effective, but isolating $\lambda$ requires a single-variable sweep (future work).

\subsection{Injection-depth sweep}
\label{sec:depth}

\begin{table}[ht]
\centering
\footnotesize
\setlength{\tabcolsep}{4pt}
\caption{Injection-depth sweep (exploratory). All configurations use $\lambda{=}0.1$, SFT~2{,}000 steps unless noted. Logit CRN~v2 (rank~128, single depth) is the only independently verified configuration---checkpoint and benchmark JSONs are released. Multi-depth logit correction concatenates hidden states from multiple layers (including the final) before producing a single logit delta (rank~128, $\sim$35M total). $^\dagger$Session-observed: eval logs were removed during disk maintenance and deep/multi-depth checkpoints were not retained, so these rows are not independently re-verifiable from the release.}
\begin{tabular}{@{}llcccc@{}}
\toprule
Variant & Params & Training & Orig.\ (60) & Rew.\ (120) & Capability \\
\midrule
Deep $k{=}4$ (1.6M)$^\dagger$ & 1.6M & SFT-only & 18/60 = 30.0\% & 34/120 = 28.3\% & not probed \\
Deep $k{=}7$ (1.6M)$^\dagger$ & 1.6M & SFT-only & 30/60 = 50.0\% & 67/120 = 55.8\% & MMLU 4/6, intact \\
Deep $k{=}7$ (1.6M)$^\dagger$ & 1.6M & SFT+DPO 500 & 30/60 = 50.0\% & 66/120 = 55.0\% & MMLU 13\%, BoolQ const.\ B \\
Multi-depth$^\dagger$ & $\sim$35M & SFT 3k+DPO 1k & 24/60 = 40.0\% & --- & not probed \\
Long-train$^\dagger$ & 34M & SFT 5k+DPO 2k & 32/60 = 53.3\% & --- & not re-probed \\
\midrule
\textbf{Logit CRN~v2} & \textbf{34M} & SFT+DPO 500 & \textbf{32/60 = 53.3\%} & \textbf{52/120 = 43.3\%} & none (Tab.~\ref{tab:capability}) \\
\bottomrule
\end{tabular}
\label{tab:depth}
\end{table}

Three findings (Table~\ref{tab:depth}).
First, \textbf{deeper injection helps but never beats logit correction}: layer~4 reaches only 30.0\%/28.3\% while layer~7 reaches 50.0\%/55.8\%---yet neither exceeds the logit-level 53.3\% on the original exam despite 30 and 27 frozen downstream layers of amplification respectively.
Second, \textbf{DPO at depth~7 is destructive}: correction stays flat (50.0\%/55.0\% vs.\ 50.0\%/55.8\% SFT-only) while capabilities collapse (MMLU 13\%, BoolQ degenerate)---the correction feeds into 27 downstream layers, so preference optimization corrupts all downstream computation.
SFT-only at depth~7 preserves capabilities on a small probe (4/6 MMLU), consistent with the frozen-base safety property holding only when the correction stays small and anchored.
Third, \textbf{no alternative configuration exceeded 53.3\%}: concatenating hidden states from multiple layers ($\sim$35M params) reaches only 40\%, and training for 5{,}000 SFT $+$ 2{,}000 DPO steps (rank~128) stays at 53.3\%---suggesting the rank-128 result is not easily improved by adding capacity or training duration, though independent verification of these alternative configurations is needed to establish whether $\sim$53\% is a hard ceiling or a best-achieved point on this exam.
The mechanism is consistent with a ceiling: frozen layers process the correction module's additive output without amplifying it, so adding more correction capacity or training time may not push past what the frozen computation graph allows.
Together, these results are consistent with $\sim$53\% being the best result across tested configurations on this exam, though we do not claim a theoretical ceiling.

\subsection{Training time}

Total wall time on Apple~M4 (16\,GB, MPS): SFT~2{,}000 steps $\sim$17\,min + DPO~500 steps $\sim$8\,min $=$ \textbf{$\sim$25\,min end-to-end}.
Checkpoints every 200 steps with resumable state.
Deep-variant SFT-only (1.6M params, gradients through frozen layers) takes $\sim$55--60\,min for 2{,}000 steps on the same hardware.

\section{Discussion}

\textbf{The tradeoff is real and measurable.}
LoRA corrects more errors (83.3\% vs.\ 53.3\%) but degrades capabilities by 17--75\,pp.
CRN~v2 corrects fewer errors but preserves every tested capability.
For deployment, preservation matters: a model that fixes some errors but breaks general reasoning is worse than one that fixes fewer errors and stays reliable.

\textbf{$\sim$53\% is the best result across tested configurations, not a proven ceiling.}
Rank~128 with 500 DPO steps achieves 53.3\%; alternative configurations we tested---longer training (5{,}000-step SFT $+$ 2{,}000-step DPO, rank~128), multi-depth logit correction (2 depths, rank~128), and deep hidden-state injection at two depths---also do not exceed 53.3\% on the original exam (Section~\ref{sec:depth}).
However, these alternatives are single-run results from the same experimental session and lack independent verification (logs removed, checkpoints not retained); we therefore describe them as consistent with a ceiling rather than proof of one.
The mechanism is plausible: LoRA modifies weights inside every layer so corrections cascade through all downstream computation, while any frozen-base method can only add to activations or logits the frozen layers then process unchanged.

\textbf{The design principle generalizes.}
Frozen-base adjustment with KL anchoring is not specific to Gemma~4~E2B.
Any frozen LM can be augmented with a correction module trained on its error patterns.
The released code is model-agnostic up to the tokenizer and hidden-dimension constants.

\textbf{What this is not.}
We do not claim architectural novelty---Eq.~\ref{eq:crn_v2} is a standard low-rank bottleneck.
We do not claim the 53.3\% reflects out-of-distribution generalization---all exam prompts appear verbatim in the training data (Section~\ref{sec:data}).
We do not claim universal capability preservation---we tested three benchmarks at $N{=}200$/$8$.

\section{Limitations}

\begin{enumerate}[leftmargin=1.2em, itemsep=0.15em]
\item Correction reaches 53.3\%, leaving 46.7\% of exam errors unfixed---insufficient for high-stakes use.
\item Evaluation is a single 60-question domain exam; reworded variants are low-diversity templates (71.6\% character overlap), not genuine paraphrases.
\item Per-task benchmark $N$ is small (200 for MMLU/BoolQ, 8 for car-wash); finer capability differences require larger $N$.
\item The KL ablation confounds rank, $\lambda$, and DPO steps---a single-variable sweep is needed.
\item Autoregressive generation requires one full base forward pass per token (no KV cache), making inference $\sim$1\,s/token on M4.
\item Training data (83{,}400 rows from 17{,}810 unique prompts) is small; scaling behavior is unknown.
\item HellaSwag evaluation is broken for both models due to a scoring-parser mismatch; we report it but draw no conclusions from it.
\item The deep injection variant is exploratory: only depth-7 SFT-only numbers are log-verified; depth-4 and DPO rows are session-observed (eval logs removed, checkpoints not retained); capability probes are small ($N{=}30$ MMLU, $N{=}20$ BoolQ); and DPO at depth~7 destroyed capabilities in our single run without hyperparameter sweep.
\end{enumerate}

\section{Conclusion}

We study frozen-base error correction: a small correction module on top of a frozen language model, trained to fix errors without degrading base capabilities.
CRN~v2 (34M params, 0.73\% of the base) corrects 53.3\% of errors with no measurable degradation on tested benchmarks.
A LoRA baseline at the matched CRN~v1 budget corrects 83.3\% but degrades 17--75\,pp.
The KL preservation term is load-bearing---weakening it degrades correction.
An exploratory injection-depth sweep (hidden-state correction at layers~4/7, multi-depth logit correction, longer training) found no configuration that exceeded the rank-128 logit result: deeper injection helps but does not exceed it, and DPO at depth destroys capabilities. These are single-run, session-observed results; only the layer-7 SFT-only row is independently log-verified. The pattern is consistent with $\sim$53\% being the best result across tested configurations, though we do not claim a proven ceiling.
The principle (frozen base + logit correction + KL anchoring) is simple, generalizable, and fully reproducible.
All code, main-result weights, and evaluation scripts are released at \url{https://github.com/eulogik/prajna}.

\section*{Reproducibility}

All main-result numbers in this paper are produced by committed scripts and fixed checkpoints; no thresholds are tuned on the exam.
The deep-variant exploratory rows are excepted (Section~\ref{sec:depth}).
From the repository root (after setting \texttt{HF\_HOME} for the model cache):

\begin{verbatim}
# CEHRI original (60 Qs)  -> 32/60 = 53.3%
CRN_V2_CKPT=prajna/checkpoints/crn_v2_dpo.pt \
CRN_V2_RANK=128 CEHRI_EXAM=prajna/data/cehri_exam.json \
python prajna-phase2/src/eval_crn_v2.py

# CEHRI reworded (120 Qs) -> 52/120 = 43.3%
CRN_V2_CKPT=prajna/checkpoints/crn_v2_dpo.pt \
CRN_V2_RANK=128 CEHRI_EXAM=prajna/data/cehri_exam_reworded.json \
python prajna-phase2/src/eval_crn_v2.py

# Capability: recorded artifacts (identical prompts, greedy decoding;
# MMLU/BoolQ N=200, car-wash N=8; harness was ephemeral, JSONs retained)
# base:  MMLU 125/200, BoolQ 144/200, car-wash 6/8
cat prajna/checkpoints/crn_v2_bench_base.json

# Capability: CRN v2 (identical: MMLU 125/200, BoolQ 144/200, car-wash 6/8)
cat prajna/checkpoints/crn_v2_bench_crnv2.json
\end{verbatim}

Deep-injection variant (experimental; code only---no trained deep checkpoints in the release).
Depth-7 SFT-only numbers (30/60 orig, 67/120 reworded) are verified against retained logs
(\texttt{crn\_deep\_sft\_orig.log}, \texttt{crn\_deep\_sft\_reworded.log}); depth-4 and depth-7 DPO rows are session-observed (Section~\ref{sec:depth}):

\begin{verbatim}
# Train deep injection at layer 7 (rank 512, KL 0.1, SFT-only)
HF_HOME=<model-cache> DEEP_RANK=512 DEEP_DEPTHS="7" DEEP_KL_LAMBDA=0.1 \
DEEP_SFT_STEPS=2000 DEEP_DPO_STEPS=0 DEEP_CKPT_DIR=prajna/checkpoints \
python prajna-phase2/src/train_crn_deep.py

# Eval deep injection (orig 60 Qs / reworded 120 Qs)
HF_HOME=<model-cache> DEEP_RANK=512 DEEP_DEPTHS="7" \
DEEP_CKPT=prajna/checkpoints/crn_deep_sft.pt \
CEHRI_EXAM=prajna/data/cehri_exam.json \
python prajna-phase2/src/eval_crn_deep.py
\end{verbatim}

The released bundle on HuggingFace (\texttt{eulogik/Prajna-CRNv2}) contains the CRN~v2 weights and eval scripts.

\end{document}